\documentclass{article}
\usepackage{spconf,amsmath,graphicx,hyperref}

\usepackage{booktabs}
\usepackage[table]{xcolor}
\usepackage{array}
\usepackage{amssymb}

\title{RefineBridge: Generative Bridge Models improve financial forecasting by Foundation Models}
\name{Anthony Bolton$^{1,\star}$\thanks{$^\star$ Equal Contribution}, Wuyang Zhou$^{1,\star}$, Zehua Chen$^{1,2,\dagger}$\thanks{$^\dagger$ Corresponding Author}, Giorgos Iacovides$^{1}$, Danilo Mandic$^{1}$}
\address{
$^1$Department of Electrical and Electronic Engineering, Imperial College London, UK\\
$^2$Department of CST, Tsinghua University, Beijing, China}
\begin{document}
\ninept
\maketitle
\begin{abstract}
Financial time series forecasting is particularly challenging for transformer-based time series foundation models (TSFMs) due to non-stationarity, heavy-tailed distributions, and high-frequency noise present in data. 
Low-rank adaptation (LoRA) has become a popular parameter-efficient method for adapting pre-trained TSFMs to downstream data domains. However, it still underperforms in financial data, as it preserves the network architecture and training objective of TSFMs rather than complementing the foundation model. 
To further enhance TSFMs, we propose a novel refinement module, \textit{RefineBridge}, built upon a tractable Schrödinger Bridge (SB) generative framework. %
Given the forecasts of TSFM as generative prior and the observed ground truths as targets, RefineBridge learns context-conditioned stochastic transport maps to improve TSFM predictions, iteratively approaching the ground-truth target from even a low-quality prior. 
Simulations on multiple financial benchmarks demonstrate that RefineBridge consistently improves the performance of state-of-the-art TSFMs across different prediction horizons.

\end{abstract}
\section{Introduction}
\label{sec:intro}

Inspired by the scaling laws observed in foundation models for natural language processing \cite{grattafiori2024llama} and computer vision \cite{dosovitskiy2020image}, Time Series Foundation Models (TSFM) \cite{chronos,moirai,timemoe, rasul2024lagllama, liutimer} have extended this scaling paradigm to time series forecasting, with models containing millions to billions of parameters. By scaling the transformer architecture \cite{vaswani2017attention} and pre-training on diverse time series datasets, TSFMs can learn universal temporal representations and have achieved state-of-the-art performance across a wide range of forecasting tasks. 

Despite the presence of financial time series in the training data of TSFMs, their prediction remains particularly challenging due to domain-specific statistical properties~\cite{fintsb2025benchmark}. In particular, financial data often exhibit near-zero autocorrelation, strong non-stationarity, and heteroskedasticity which shifts across market regimes. As a result, TSFMs tend to default to mean-reverting forecasts and output the average of the context window irrespective of market conditions. This leads to predictions that miss higher-frequency variations, which frequently carry tradable signals.

\begin{figure}[h]
  \centering
  \includegraphics[width=\columnwidth]{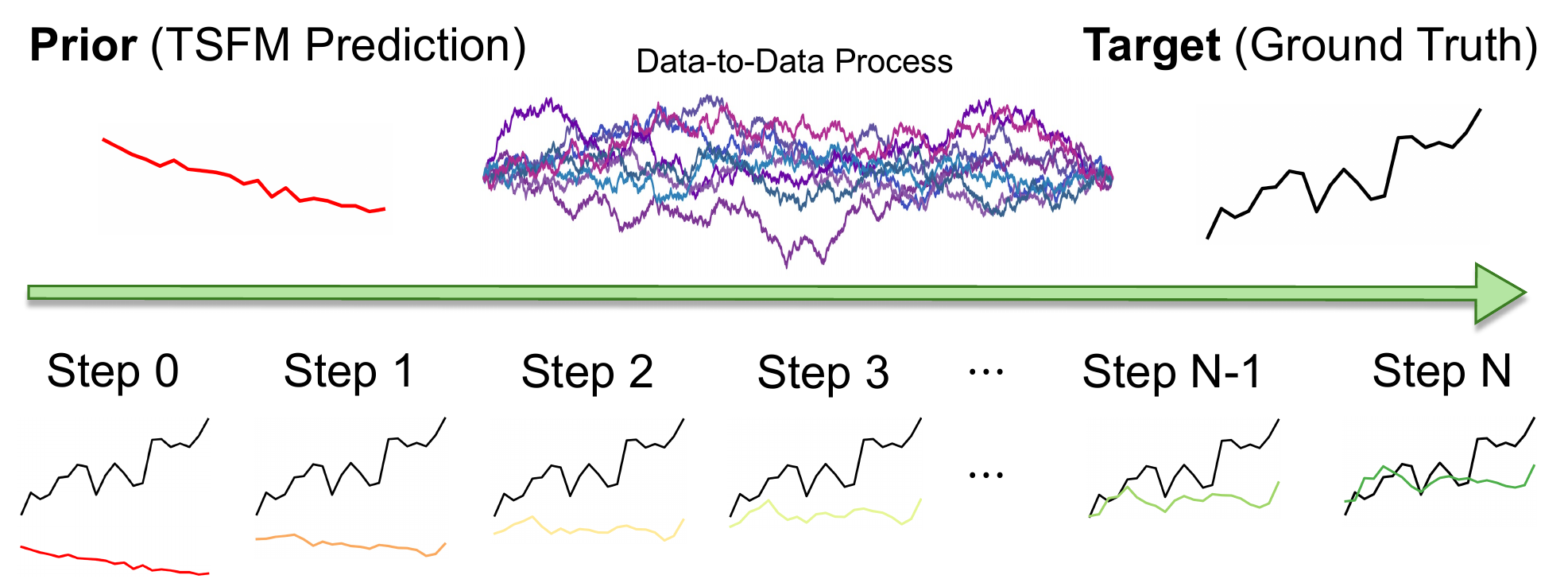}
  \caption{RefineBridge learns an optimal transport map from TSFM predictions to the ground truth via Schrödinger Bridge. At inference time, RefineBridge progressively refines the TSFM forecasts.} 
  \label{fig:arch}
  \vspace{-0.4cm}
\end{figure}

Low-Rank Adaptation (LoRA)~\cite{lora} has emerged as a popular parameter-efficient method for adapting TSFMs to downstream data domains~\cite{gupta2024beyond}. LoRA-based adapters introduce low-rank update matrices with fewer trainable parameters than the original foundation model, enabling parameter-efficient fine-tuning. However, when applied to financial data, LoRA is prone to overfitting and can even degrade TSFM performance (see Table~\ref{tab1}). 
Specifically, LoRA typically adopts the same network architecture and objective function as the original model, rather than introducing a complementary objective to enhance predictions, inherently limiting its performance.

To address these issues, we propose \textit{RefineBridge}, a novel paradigm for improving the performance of TSFMs on financial time series forecasting. 
In contrast to LoRA, which functions as a fine-tuning technique, RefineBridge operates as an independent post-processing module. Specifically, it treats TSFM forecasts as the prior distribution and iteratively refines them through Schrödinger Bridge optimal transport~\cite{bridgetts,wang2025framebridge,audiolbm,voicebridge}, conditioned on the context window. 
As a probabilistic generative model, RefineBridge introduces a complementary training objective which captures the optimal stochastic transport maps from the prior to the target distribution. By using the coarse predictions by TSFMs as a prior, RefineBridge learns to model fine-grained details, which are crucial in financial time series forecasting.
During the sampling stage, RefineBridge generates a refinement trajectory which aligns TSFM forecasts with the unique statistical properties of downstream financial data, all while keeping the original TSFM parameters unchanged, as illustrated in Fig.~\ref{fig:arch}.
The main contributions of this work are as follows.

\begin{itemize}
    \item We design \textit{RefineBridge}, a Schrödinger Bridge-based module to enhance TSFMs in financial time series forecasting.  
    \item Different from LoRA, which fine-tunes TSFMs on downstream data domains, RefineBridge is an independent module and introduces a different training objective to utilize the predictions of TSFMs and capture the target distribution.  
    \item Extensive experiments on financial time series demonstrate that RefineBridge consistently improves the performance of TSFMs and outperforms LoRA across multiple downstream finance datasets and forecast horizons.  
\end{itemize}

\section{Related Work}
\noindent \textbf{Time Series Foundation Models.} Recent advancements in deep learning have led to the development of various TSFMs~\cite{chronos,moirai,timemoe,rasul2024lagllama,liutimer}, with Chronos \cite{chronos}, Moirai \cite{moirai}, and Time-MoE \cite{timemoe} representing three distinct architectures with different training objective functions. Chronos~\cite{chronos} adapts the T5~\cite{t5} architecture by discretizing continuous time series into tokens and treating forecasting as a sequence-to-sequence task, where future tokens are generated auto-regressively. Moirai~\cite{moirai} instead uses mixture distributions and a masked encoder architecture with continuous embeddings. At a larger scale, Time-MoE~\cite{timemoe} is the first TSFM to reach the billion-parameter level through a sparse mixture-of-experts architecture. \par 

\noindent \textbf{Low-rank Adaptation.} Full fine-tuning of TSFMs for specific downstream datasets is computationally expensive due to their large parameter count. To address this, LoRA-based adapters~\cite{lora,kopiczko2024vera,gu2025tera} introduce low-rank weight update matrices to greatly reduce the number of trainable parameters during fine-tuning. However, in practice, LoRA struggles when applied to financial data, as shown in Table~\ref{tab1}.

\section{RefineBridge}

\begin{figure*}[t] 
  \centering
  \includegraphics[width=\textwidth]{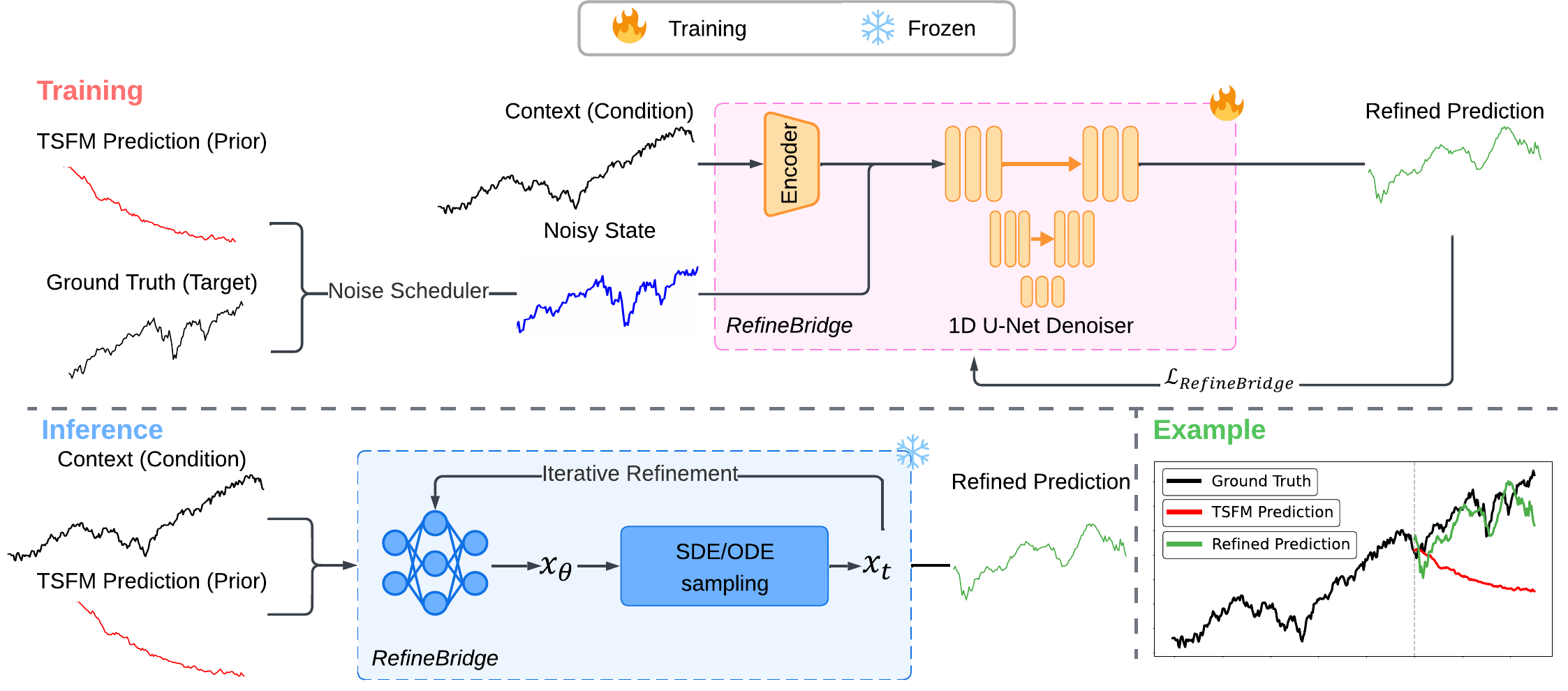}
  \caption{The proposed RefineBridge model learns how to refine the predictions of TSFMs towards ground truth conditioned on the context window during training. During inference, the trained model is frozen and uses SDE or ODE sampling to iteratively refine the predictions of TSFMs conditioned on the context window. The bottom right corner shows an example of a comparison between the original TSFM prediction and the refined prediction through RefineBridge.}
  \label{fig:wide_diagram}
\end{figure*}

Unlike diffusion models which map noise to data, Schrödinger Bridge learns data-to-data transport maps by computing the most probable stochastic evolution between two probability distributions under a reference diffusion process~\cite{sb_fbsde, diffusion_schrodinger_bridge}. This makes SB well-suited for refining TSFM predictions, which can be viewed as neural priors learned by the internal representation of TSFMs, as illustrated in Fig. \ref{fig:arch}. Even though TSFMs often struggle with financial time series, their forecasts still contain a coarse estimate (the neural prior) of the prediction target. As highlighted by \cite{wang2025framebridge}, compared to noise-to-data diffusion models, SB-based models can capture the target distribution from a clean prior, making it possible to build post-processing refinement models for pretrained foundation models.

\subsection{Bridging TSFM Predictions and Ground Truth}
To refine TSFM forecasts on downstream datasets, we formulate forecast refinement as a conditional Schrödinger Bridge problem between TSFM predictions and the ground truth. Given a historical context window $\mathbf{c} \in \mathbb{R}^{C}$ of $C$ time steps, a TSFM generates prediction, $\mathbf{x}_T = \mathcal{F}(\mathbf{c}) \in \mathbb{R}^{H}$, for the prediction horizon $H$. Our objective is to learn a stochastic refinement process that transports $\mathbf{x}_T \sim p_{\text{prior}}$ to the ground truth $ \mathbf{x}_0 \sim p_{\text{data}}$ by minimizing the Kullback-Leibler divergence, $D_{\text{KL}}$, between a path measure, $p$, and reference path measure, $p_{\text{ref}}$, as 
\begin{equation}
\min_{p \in \mathcal{P}_{[0,T]}} D_{\text{KL}}(p \| p_{\text{ref}}) \quad \text{s.t.} \quad p_0 = p_{\text{data}}, \, p_T = p_{\text{prior}|c},
\end{equation}
where $\mathcal{P}_{[0,T]}$ denotes the collection of all path measures over the time interval $[0,T]$, and the prior distribution $p_{\text{prior}|c}$ is conditioned on the context.

Following prior work \cite{bridgetts, wang2025framebridge, li2025bridge}, we adopt a tractable solution for paired data $( \mathbf{x}_0, \mathbf{x}_{T})$. We set the reference stochastic differential equation (SDE) to have zero drift ($f(t)=0$) and a time-varying diffusion defined as $d \mathbf{x}_t = g(t)d\mathbf{w}_t$, where $ g^2(t) = \beta_0 + t(\beta_1 - \beta_0)$. Here, $f : \mathbb{R}^H \times [0, T] \to \mathbb{R}^H$ is the vector-valued drift term, $g : [0, T] \to \mathbb{R}$ is scalar-valued diffusion coefficient, and $\mathbf{w}_t \in \mathbb{R}^H$  denotes a standard Wiener process. This SDE defines the continuous-time diffusion process upon which score-based generative models (SGMs)~\cite{sgm,unicardio,audioldm,diffgap,freeaudio,controlaudio} are built.

In the general SB framework with linear drift, the scaling coefficients are given by $\alpha_t = e^{\int_0^t f(\tau)d\tau}$ and $\bar{\alpha}_t = e^{-\int_t^1 f(\tau)d\tau}$. We employ the \textit{Bridge-gmax} scheduler proposed in \cite{bridgetts}, to simplify these to $\alpha_t = \bar{\alpha}_t = 1$ under $f(t) = 0$. The corresponding variance schedules are $\sigma_t^2 = \int_0^t g^2(\tau)d\tau = \beta_0 t + \frac{1}{2}(\beta_1 - \beta_0)t^2$ and $ \bar{\sigma}_t^2 = \sigma_1^2 - \sigma_t^2.$ The resulting bridge marginal distribution is Gaussian
\begin{align}
p_t( \mathbf{x}_t| \mathbf{x}_T, \mathbf{x}_0, \mathbf{c}) = 
\mathcal{N}\left(\frac{\bar{\sigma}^2_t \mathbf{x}_T+ \sigma^2_t \mathbf{x}_{0}}{\sigma^2_1}, 
\frac{\sigma^2_t \bar{\sigma}^2_t}{\sigma^2_1}\mathbf{I}_H\right).
\label{eq:bridge_marginal}
\end{align}

Given a trained RefineBridge model $x_\theta( \mathbf{x}_t, t, \mathbf{x}_{T}, \mathbf{z}_c)$, used to refine the prior $\mathbf{x}_T$ conditioned on the encoded context $\mathbf{z}_c = \mathcal{E}(\mathbf{c})$, the conditional refinement process can be expressed as 
\begin{equation}
d \mathbf{x}_t = g^2(t)\frac{ \mathbf{x}_t -x_\theta( \mathbf{x}_t, t, \mathbf{x}_{T}, \mathbf{z}_c)}{\sigma_t^2}dt + g(t)d\bar{\mathbf{w}}_t,
\end{equation}
for \textbf{stochastic refinement} and as 
\begin{equation}
d \mathbf{x}_t = \frac{1}{2}g^2(t)\left[\frac{ \mathbf{x}_t -x_\theta( \mathbf{x}_t, t, \mathbf{x}_{T}, \mathbf{z}_c)}{\sigma_t^2} - \frac{ \mathbf{x}_t - \mathbf{x}_{T}}{\bar{\sigma}_t^2}\right]dt.
\end{equation}
for \textbf{deterministic refinement}.
\subsection{Complementary Training Objective}
When LoRA~\cite{lora} is applied to a TSFM for downstream adaptation, it typically retains the training objective of the base model, thereby potentially inheriting its limitations. This includes maximizing token-level log-likelihood in Chronos~\cite{chronos}, optimizing a mixture distribution log-likelihood in Moirai~\cite{moirai}, and minimizing a multi-resolution autoregressive Huber loss in Time-MoE~\cite{timemoe}. As these models are pretrained for time series forecasting, their loss functions are directly reused during LoRA during fine-tuning on downstream datasets. Consequently, LoRA inherits the same network architecture, training objectives, and potential drawbacks of the original TSFM. Take Time-MOE \cite{timemoe} as an example. Its multi-resolution loss function is defined as 
\begin{equation}
\mathcal{L} = \frac{1}{P}\sum_{j=1}^{P} \mathcal{L}_{\text{ar}} + \alpha\mathcal{L}_{\text{aux}},
\end{equation}
where $\mathcal{L}_{\text{ar}}$ denotes the Huber loss \cite{huberloss}, $P$ is the number of different prediction resolutions, $\alpha$ is a scaling factor, and $\mathcal{L}_{\text{aux}}$ is an auxiliary loss for routing in the mixture-of-experts (MoE) layers \cite{dai2024deepseekmoe, riquelme2021scaling}. When fine-tuning Time-MoE on a downstream dataset with LoRA, its training objective must continue to balance the Huber loss and auxiliary MoE loss. Furthermore, the pretraining objectives of many TSFMs are often designed to learn generalized patterns through coarse predictions \cite{chronos}. While beneficial for generalization across multiple domains, this can be a significant drawback for financial applications, where subtle predictive signals are crucial for accurate forecasting under changing market regimes.

RefineBridge differs from LoRA-based approaches by introducing a different training objective. Namely it minimizes the conditional denoising loss
\begin{equation}
\begin{split}
&\mathcal{L} = \mathbb{E}_{(\mathbf{z}_c, \mathbf{x}_0) \sim p_{\text{data}}, \mathbf{z}_c =  \mathcal{E}(\mathbf{c})} \mathbb{E}_{t} \left[\|x_\theta( \mathbf{x}_t, t, \mathbf{x}_{T}, \mathbf{z}_c) - \mathbf{x}_0\|_2^2\right],
\end{split}
\end{equation}
where $ \mathbf{x}_t$ is sampled from the bridge marginal distribution as described in  (\ref{eq:bridge_marginal}). 
This training objective enables bridge models to learn time-dependent score functions, allowing them to gradually reconstruct the target distribution from the predefined prior distribution,~\textit{e.g.}, the forecasts of TSFMs, in post-processing.
This difference allows our proposed RefineBridge to bypass a constraint of LoRA, where the same neural architecture is used. In summary, the time-dependent conditional denoising loss of the probabilistic generative framework is exploited as a complementary training objective of the TSFM to improve the final time series predictions.

\begin{table*}[t]
\centering
\caption{Forecasting performance across five horizons for Chronos, Moirai, and Time-MoE. Our RefineBridge is denoted in Bold. The prediction horizons are $5, 10, 21, 63,$ and $126$. \textbf{Best} performance is denoted in bold, and \underline{second best} is underlined.}
\setlength{\tabcolsep}{1.7pt} 
\scriptsize
\resizebox{\textwidth}{!}{%
\begin{tabular}{cc|cc|cc|cc!{\vrule width 0.8pt}cc|cc|cc!{\vrule width 0.8pt}cc|cc|cc}
\toprule
\multicolumn{2}{c}{} 
  & \multicolumn{6}{c!{\vrule width 0.8pt}}{\textsc{Chronos}$_{\text{large}}$}
  & \multicolumn{6}{c!{\vrule width 0.8pt}}{\textsc{Moirai}$_{\text{large}}$}
  & \multicolumn{6}{c}{\textsc{Time-MoE}$_{\text{base}}$} \\
\cmidrule(lr){3-8}\cmidrule(lr){9-14}\cmidrule(lr){15-20}
\multicolumn{2}{c}{\textbf{Metrics}}
  & \multicolumn{2}{c}{Original} & \multicolumn{2}{c}{LoRA} & \multicolumn{2}{c!{\vrule width 0.8pt}}{\textbf{RefineBridge}}
  & \multicolumn{2}{c}{Original} & \multicolumn{2}{c}{LoRA} & \multicolumn{2}{c!{\vrule width 0.8pt}}{\textbf{RefineBridge}}
  & \multicolumn{2}{c}{Original} & \multicolumn{2}{c}{LoRA} & \multicolumn{2}{c}{\textbf{RefineBridge}} \\
\cmidrule(lr){3-4}\cmidrule(lr){5-6}\cmidrule(lr){7-8}
\cmidrule(lr){9-10}\cmidrule(lr){11-12}\cmidrule(lr){13-14}
\cmidrule(lr){15-16}\cmidrule(lr){17-18}\cmidrule(lr){19-20}
\textbf{Asset} & H
  & MSE & MAE & MSE & MAE & MSE & MAE
  & MSE & MAE & MSE & MAE & MSE & MAE
  & MSE & MAE & MSE & MAE & MSE & MAE \\
\midrule
  & 5   & \underline{0.009} & \underline{0.075} & 0.011 & 0.078 & \textbf{0.008} & \textbf{0.067}
            & \underline{0.016} & \underline{0.095} & 0.023 & 0.110 & \textbf{0.008} & \textbf{0.069}
            & 0.012 & 0.088 & \underline{0.011} & \underline{0.079} & \textbf{0.009} & \textbf{0.073} \\
  & 10  & \underline{0.016} & \underline{0.101} & 0.017 & 0.099 & \textbf{0.014} & \textbf{0.092}
            & \underline{0.021} & \underline{0.111} & 0.026 & 0.122 & \textbf{0.017} & \textbf{0.098}
            & 0.022 & 0.121 & \textbf{0.017} & \textbf{0.103} & \underline{0.018} & \underline{0.105} \\
{\textbf{S\&P500}}  & 21  & \underline{0.157} & \underline{0.324} & 1.894 & 1.339 & \textbf{0.105} & \textbf{0.260}
            & \underline{0.115} & \underline{0.267} & 0.898 & 0.329 & \textbf{0.103} & \textbf{0.251}
            & \underline{0.357} & \underline{0.491} & 1.330 & 1.098 & \textbf{0.279} & \textbf{0.426} \\
  & 63  & \underline{1.164} & \underline{0.894} & 3.308 & 1.025 & \textbf{0.567} & \textbf{0.601}
            & \underline{0.388} & \underline{0.487} & 3.972 & 1.906 & \textbf{0.297} & \textbf{0.427}
            & 2.542 & 1.391 & \underline{2.293} & \underline{1.389} & \textbf{1.328} & \textbf{1.007} \\
  & 126 & \underline{5.536} & \underline{2.026} & 6.493 & 2.457 & \textbf{2.102} & \textbf{1.207}
            & \underline{1.566} & \underline{0.996} & 5.550 & 2.188 & \textbf{0.764} & \textbf{0.691}
            & 9.257 & 2.761 & \textbf{2.511} & \textbf{1.326} & \underline{3.338} & \underline{1.603} \\
\midrule
  & 5   & \underline{0.175} & \underline{0.319} & 0.252 & 0.373 & \textbf{0.146} & \textbf{0.293}
            & 0.351 & 0.447 & \underline{0.291} & \underline{0.412} & \textbf{0.178} & \textbf{0.321}
            & \underline{0.198} & 0.343 & 0.202 & \underline{0.341} & \textbf{0.146} & \textbf{0.295} \\
  & 10  & \underline{0.278} & \underline{0.410} & 0.445 & 0.497 & \textbf{0.232} & \textbf{0.373}
            & 0.554 & 0.574 & \underline{0.458} & \underline{0.522} & \textbf{0.280} & \textbf{0.408}
            & \underline{0.313} & \underline{0.438} & 0.319 & 0.441 & \textbf{0.237} & \textbf{0.378} \\
{\textbf{WTI}}  & 21  & \underline{0.407} & \underline{0.503} & 0.586 & 0.627 & \textbf{0.335} & \textbf{0.469}
            & \underline{0.409} & \underline{0.502} & 1.338 & 0.965 & \textbf{0.325} & \textbf{0.458}
            & \underline{0.370} & \underline{0.478} & 0.395 & 0.501 & \textbf{0.316} & \textbf{0.457} \\
  & 63  & 0.907 & 0.771 & \underline{0.819} & \textbf{0.715} & \textbf{0.774} & \underline{0.721}
            & \underline{1.338} & \underline{0.943} & 3.319 & 1.505 & \textbf{0.891} & \textbf{0.769}
            & 2.092 & 1.223 & \underline{1.897} & \underline{1.134} & \textbf{0.928} & \textbf{0.798} \\
  & 126 & \underline{1.100} & \underline{0.844} & 1.176 & 0.876 & \textbf{0.844} & \textbf{0.742}
            & \underline{1.813} & \underline{1.047} & 2.174 & 1.190 & \textbf{1.072} & \textbf{0.863}
            & 4.010 & \underline{1.523} & \underline{3.630} & 1.569 & \textbf{1.160} & \textbf{0.904} \\
\midrule

  & 5   & \textbf{0.295} & \textbf{0.418} & \underline{0.315} & \underline{0.432} & 0.333 & 0.443
            & 0.627 & 0.629 & \underline{0.596} & \underline{0.619} & \textbf{0.327} & \textbf{0.447}
            & 0.419 & 0.500 & \underline{0.372} & \underline{0.476} & \textbf{0.339} & \textbf{0.453} \\
  & 10  & \underline{0.555} & \underline{0.584} & 0.631 & 0.593 & \textbf{0.509} & \textbf{0.557}
            & 1.093 & 0.845 & \underline{0.917} & \underline{0.779} & \textbf{0.567} & \textbf{0.599}
            & 0.786 & 0.696 & \underline{0.756} & \underline{0.685} & \textbf{0.566} & \textbf{0.597} \\
 {\textbf{EURUSD}} & 21  & \underline{1.017} & \underline{0.802} & 1.347 & 0.896 & \textbf{0.888} & \textbf{0.756}
            & \underline{1.052} & \underline{0.801} & 1.733 & 1.030 & \textbf{0.897} & \textbf{0.748}
            & 1.224 & 0.891 & \textbf{1.018} & \textbf{0.773} & \underline{1.043} & \underline{0.806} \\
  & 63  & \underline{3.381} & \underline{1.404} & 8.423 & 2.119 & \textbf{1.596} & \textbf{0.966}
            & \underline{3.263} & \underline{1.298} & 4.872 & 1.714 & \textbf{1.730} & \textbf{1.008}
            & \underline{4.626} & \underline{1.614} & 4.683 & 1.801 & \textbf{1.951} & \textbf{1.047} \\
  & 126 & \underline{3.328} & \underline{1.384} & 4.833 & 1.736 & \textbf{1.702} & \textbf{1.004}
            & \underline{2.621} & \underline{1.267} & 6.181 & 2.066 & \textbf{1.272} & \textbf{0.852}
            & \underline{2.956} & \underline{1.284} & 6.405 & 2.067 & \textbf{1.458} & \textbf{0.953} \\
\bottomrule
\addlinespace[0.25em]%
\multicolumn{2}{c}{\textbf{1\textsuperscript{st} Count}}
  & \multicolumn{2}{c}{\underline{2}}  
  & \multicolumn{2}{c}{1}  
  & \multicolumn{2}{c!{\vrule width 0.8pt}}{\textbf{27}}  
  & \multicolumn{2}{c}{0}  
  & \multicolumn{2}{c}{0}  
  & \multicolumn{2}{c!{\vrule width 0.8pt}}{\textbf{30}}  
  & \multicolumn{2}{c}{0}  
  & \multicolumn{2}{c}{\underline{6}}  
  & \multicolumn{2}{c}{\textbf{24}}  
\\
\bottomrule
\end{tabular}%
\label{tab1}
}
\vspace{-0.35cm}
\end{table*}

\subsection{Prior-to-Target: Iterative Refinement for TSFM}
During inference, we refine a TSFM prediction $\mathbf{x}_T$ given encoded context $\mathbf{z}_c$, as shown in Fig. \ref{fig:wide_diagram}. The refinement is performed via either deterministic sampling using ordinary differential equation (ODE) or stochastic sampling using SDE, both discretized with Euler steps. Given an initial TSFM prediction $\mathbf{x}_s = \mathbf{x}_{T}$ at $s = 1$, the first-order discretization of the ODE sampler is
\begin{equation}
\begin{split}
\mathbf{x}_t &= \frac{\sigma_t \bar{\sigma}_t}{\sigma_s \bar{\sigma}_s} \mathbf{x}_s + \frac{1}{\sigma_1^2}\left( \sigma_t^2 - \frac{\sigma_s \sigma_t \bar{\sigma}_t}{\bar{\sigma}_s} \right) \mathbf{x}_T\\
&\quad \quad+ \frac{1}{\sigma_1^2}  \left( \bar{\sigma}_t^2 - \frac{\bar{\sigma}_s \sigma_t \bar{\sigma}_t}{\sigma_s} \right) x_\theta(\mathbf{x}_s, s, \mathbf{x}_{T}, \mathbf{z}_c),
\end{split}
\end{equation}
while the first-order discretization of the SDE sampler is
\begin{equation}
\begin{split}
\mathbf{x}_t &= \frac{\sigma_t^2}{\sigma_s^2}\mathbf{x}_s + \left(1 - \frac{\sigma_t^2}{\sigma_s^2}\right)x_\theta(\mathbf{x}_s, s, \mathbf{x}_{T}, \mathbf{z}_c) \\
&\quad \quad + \sigma_t\sqrt{1 - \frac{\sigma_t^2}{\sigma_s^2}}\boldsymbol{\epsilon}_\tau,
\end{split}
\end{equation}
where $\boldsymbol{\epsilon}_\tau \sim \mathcal{N}(0, \tau^{-1}\mathbf{I}_H)$, with temperature $\tau$ controlling the stochasticity. 

Empirically, we find that, for short prediction horizons ($H \in \{5, 10\}$) where TSFM predictions remain closer to the ground truth and retain the informative prior, ODE sampling with fewer denoising steps ($1–10$) usually achieves the best performance. For longer horizons ($H \in \{21, 63, 126\}$) where TSFM predictions can deviate substantially from the ground truth and require more refinements, SDE sampling with temperature tuning ($\tau \in [0.01, 2.0]$) over more denoising steps ($10-1000$) achieves the best refining performance.

\subsection{Network Architecture}
RefineBridge trains a context encoder and a 1D U-Net denoiser in an end-to-end fashion. For the context encoder, we employ the DLinear~\cite{dlinear} model to decompose the context into its trend and seasonal components. The encoder learns the latent representation of the context window, $\mathbf{z}_c=\mathcal{E}(\mathbf{c})\in \mathbb{R}^{H \times d}$.

In the 1D-UNet, diffusion time $t$ is encoded with sinusoidal position embeddings while Conv1D residual blocks are used with GroupNorm and SiLU for data standardisation. The 1D U-Net denoiser takes the concatenated input $[\mathbf{x}_{t}, \mathbf{x}_T, \mathbf{z}_c] \in \mathbb{R}^{H \times (d+2)}$ and iteratively refines it to obtain the target $\mathbf{x}_0$. 

\section{EXPERIMENTS}
\label{sec:pagestyle}
\subsection{Datasets}
We evaluated RefineBridge on three distinct financial assets, the S\&P 500 index (equity market), WTI crude oil (commodities), and the EUR/USD exchange rate (forex). We sourced 20 years of daily data (2005-01-01 to 2025-01-01), yielding approximately 5,040 trading days per asset. 
For each asset, the data were partitioned temporally into train/validation/test (80\%/10\%/10\%) splits. Training samples were generated using a sliding window of step size 1. Each sample consisted of a triplet $\{\mathbf{x}_T,\mathbf{c}, \mathbf{x}_0 \}$. We adopted a context window length of $C=21$ trading days for short forecasting horizons ($H \in \{5, 10\}$) and $C=252$ trading days for longer forecasting horizons ($H \in \{21, 63, 126\}$). Given that financial time series exhibit heavy tails and frequent outliers which violate Gaussian assumptions, we followed~\cite{li2024master} and applied median–MAD robust normalization for all time series.
\subsection{Experimental Settings}
\textbf{Evaluation Metrics.} Table \ref{tab1} reports the average results over 5 independent runs using the Mean Squared Error (MSE) and Mean Absolute Error (MAE) metrics on the test sets. A distinct RefineBridge model was trained for each asset and was used to refine the predictions of TSFM given the context window. Metrics were then computed using the refined predictions and ground truth $\mathbf{x}_0$.\\
\textbf{Baseline Methods.} We tested RefineBridge on three distinct state-of-the-art TSFMs: \textsc{Chronos}$_{\text{large}}$ (710M)~\cite{chronos}, \textsc{Moirai}$_{\text{large}}$ (311M)~\cite{moirai}, and \textsc{Time-MoE}$_{\text{base}}$ (113M)~\cite{timemoe}. 

We also compared RefineBridge with LoRA, where separate LoRA adapters were trained for each asset, with rank $r=16$, to ensure fair comparisons. \\
\textbf{Implementation Details.} All models were trained with the AdamW optimizer ($lr = 10^{-3}$, $\epsilon=10^{-8}$) and a batch size of 512. For short prediction horizons ($H \leq 10$), a noise schedule of $\beta_0=0.01, \beta_1=50$ was used for rapid refinement, while for long horizons ($H \geq$ 21) we used $\beta_0=0.0001, \beta_1=0.02$ for finer refinements. Notably, each RefineBridge model consisted of only $2.3$M parameters, which was less than the number of parameters used in the LoRA adapters (7.6M in \textsc{Chronos}, 2.4M in \textsc{Moirai}, and 2.7M in \textsc{Time-MoE}).

\subsection{Results}
RefineBridge consistently improved the forecasts of state-of-the-art TSFMs across models, assets, and prediction horizons. As shown in Table \ref{tab1}, RefineBridge achieves the best performance in 81 out of 90 experimental configurations, with MSE reductions ranging from 11\% to 71\% across different scenarios. Overall, LoRA showed minimal improvements and often degraded base model performance (failing in 29 out of 45 configurations) due to the challenging nature of financial data, whereas RefineBridge consistently enhanced predictions for all assets and horizons, except for EURUSD with $H=5$. The performance advantages of RefineBridge are even more pronounced at longer horizons. For example, RefineBridge reduced the MSE of Chronos on S\&P 500 by 11\% at prediction horizon of 5 days, and by 62\% at a prediction horizon of 126 days. This is due to the auto-regressive nature of TSFMs, causing error accumulation at longer prediction horizons, which allows for greater improvement during refinement. \newline

\noindent \textbf{Remark 1.} A distinguishing feature of RefineBridge is that, unlike in other Schrödinger Bridge models \cite{bridgetts, wang2025framebridge, li2025bridge} where the prior may be directly part of or very close to the ground truth target, RefineBridge refines less informative or even false (low-quality) priors from TSFMs and still manages to improve the forecasting performance. This is shown in Fig. \ref{fig:arch} and Fig. \ref{fig:wide_diagram}, where RefineBridge successfully corrects a prior trending in a different direction from the ground-truth. Such performance demonstrates the robustness of bridge models and lays a solid foundation for utilising them as a post-processing module of TSFMs.

\section{CONCLUSION}
We have introduced RefineBridge, a novel framework for improving the forecast performances of TSFMs in downstream financial time series through the Schrödinger Bridge optimal mass transport strategy. Different from LoRA-based models, RefineBridge does not directly update the learned weight representations of the TSFM, but utilizes the TSFM predictions as neural priors. This allows the proposed model to introduce a complementary training objective for iteratively refining the predictions of TSFMs. The RefineBridge has demonstrated consistent performance improvements over different TSFMs, financial assets, and prediction horizons while also outperforming the popular LoRA method.

\vfill\pagebreak

\bibliographystyle{IEEEbib}
\bibliography{strings,refs}

\end{document}